%% file: acl_latex.tex
\documentclass[11pt]{article}

\usepackage[preprint]{acl}

\usepackage{times}
\usepackage{latexsym}

\usepackage[T1]{fontenc}

\usepackage[utf8]{inputenc}

\usepackage{microtype}

\usepackage{inconsolata}

\usepackage{graphicx}

\usepackage{amsmath}
\usepackage{amssymb}
\usepackage{mathtools}
\usepackage{amsthm}

\usepackage{multirow}
\usepackage[table]{xcolor}
\usepackage{caption}
\usepackage{microtype}
\usepackage{graphicx}
\usepackage{subcaption}
\usepackage{booktabs} 

\usepackage{algorithm}
\usepackage{algpseudocode}
\usepackage{xcolor}
\usepackage{amsmath}
\usepackage{amssymb}

\title{Learning When Not to Act: Mitigating Tool Abuse in Agentic Reinforcement Learning}

\author{
 \textbf{Liuji Chen\textsuperscript{1,2,}\thanks{Equal contribution.}\thanks{This work was completed during internship at ByteDance.}},
 \textbf{Dianxing Tang\textsuperscript{3,}\footnotemark[1]},
 \textbf{Xing Shi\textsuperscript{2}},
 \textbf{Dingshuo Chen\textsuperscript{1}},
\\
 \textbf{Qiang Liu\textsuperscript{1}},
 \textbf{Shu Wu\textsuperscript{1}},
 \textbf{Liang Wang\textsuperscript{1}},
\\
\\
 \textsuperscript{1}NLPR, Institute of Automation, Chinese Academy of Sciences,
\\
 \textsuperscript{2}ByteDance,
 \textsuperscript{3}Zhejiang University,
\\
\\
 \small{
   \textbf{Correspondence:} \href{mailto:chenliuji2023@ia.ac.cn}{chenliuji2023@ia.ac.cn}
 }
}

\begin{document}
\maketitle
\begin{abstract}
Agentic reinforcement learning can induce \emph{tool abuse}, where models overuse external tools even for queries solvable by internal reasoning. Existing approaches mitigate this issue with uniform tool-use penalties or hard limits, which reduce tool frequency but may also suppress useful tool-assisted exploration. We propose \textbf{EAPO}, an Efficient Agentic Policy Optimization framework that learns selective tool use. EAPO introduces tool-free trajectories into each rollout group, applies difficulty-aware reward shaping to penalize redundant tool calls mainly on easier queries, and uses confidence-aware token reweighting to improve policy learning. Across nine mathematical and knowledge-intensive reasoning benchmarks, EAPO consistently improves the accuracy--efficiency trade-off on Qwen2.5-3B, Qwen2.5-7B, and Llama3.1-8B. Compared with GRPO, EAPO improves average performance by 10.45\%, 7.27\%, and 9.69\%, while reducing average tool calls by 18.33\%, 18.33\%, and 24.59\%, respectively. These results show that agents can learn \emph{when not to use tools} without compromising tool-integrated reasoning.
\end{abstract}

\input{Sections/1-introduce}
\input{Sections/2-preliminary}

\input{Sections/3-method}
\input{Sections/4-experiments}

\section{Conclusion}
We introduced EAPO, a reinforcement learning framework for mitigating tool abuse in language agents. By combining tool-free rollouts, difficulty-aware reward shaping, and confidence-aware token reweighting, EAPO helps models decide when tools are beneficial and when direct reasoning is sufficient. Experiments on mathematical and knowledge-intensive benchmarks show that EAPO improves the accuracy--efficiency trade-off across model families and scales. These results underscore the need to optimize language agents for both task success and efficient tool use.

\section*{Limitations}

In this work, we propose EAPO, a multi-tool agentic reinforcement learning algorithm that jointly optimizes both effectiveness and efficiency. While EAPO achieves strong performance across a wide range of tasks, existing benchmarks predominantly focus on hard problems, emphasizing the model’s ability to solve complex queries. However, assessing whether a model suffers from severe tool abuse requires a different evaluation paradigm. Specifically, it is necessary to construct an easy-to-hard dataset that simultaneously evaluates both the effectiveness and the efficiency of the model.

\section*{Ethical Considerations}
In this work, all training data and frameworks used by EAPO comply with their respective licenses and do not infringe on any privacy; they are used solely for academic research purposes.

\bibliography{custom}
\input{Sections/5-appendix}

\end{document}

%% file: Sections/1-introduce.tex
\section{Introduction}

\begin{figure}[ht]
  \vskip 0.2in
  \begin{center}
    \centerline{\includegraphics[width=\columnwidth]{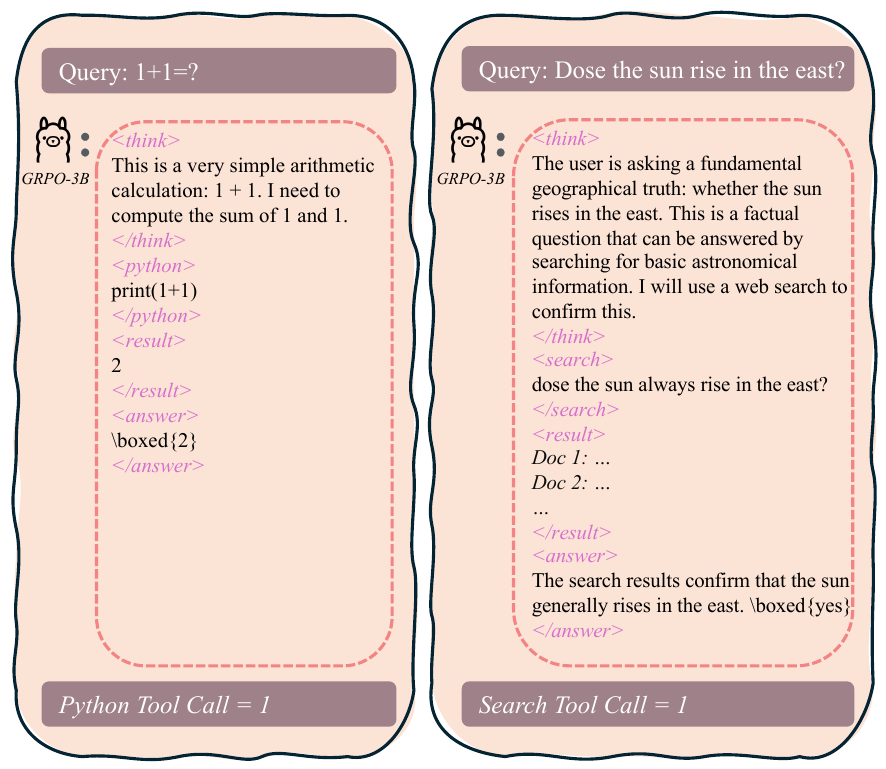}}
    \caption{
      An illustrative example of tool abuse: after agentic RL training, the Qwen-3B model becomes overly reliant on tools, invoking them even for trivial questions.
    }
    \label{Fig1: Example of tool abuse}
  \end{center}
  \vspace{-20pt}
\end{figure}

Recent advances in large language models (LLMs) have motivated growing interest in reinforcement learning (RL) for tool-augmented agentic reasoning. By allowing models to interact with external tools such as search engines or code interpreters, these methods can substantially improve problem-solving ability on complex tasks \cite{zhang2026landscapeagenticreinforcementlearning, wei2025autotir, jin2025searchr1, xue2025simpletir, li2025torlscalingtoolintegratedrl, dong2025arpo, dong2025aepo}. Compared with conventional retrieval-augmented generation pipelines, agentic frameworks enable models to decide when and how to invoke tools, making them more flexible in handling multi-step reasoning problems.

However, this flexibility also introduces a practical challenge that remains underexplored: \textbf{tool abuse}. Since existing agentic RL objectives mainly reward task success in tool-available environments, models may gradually learn to treat tool invocation as a default reasoning strategy. As shown in Figure~\ref{Fig1: Example of tool abuse}, a model trained with external tools may invoke Python or search tools even for elementary arithmetic or simple commonsense questions that can be solved directly through internal reasoning. Although such behavior may still produce correct answers, it introduces unnecessary computation, latency, and deployment cost, especially when external services are billed per use.

A seemingly straightforward solution is to penalize tool usage, as in OTC \cite{wang2025actingreasoningmoreteaching}. However, simply discouraging frequent tool calls can be problematic. Tool-integrated reasoning is intended to enable models to leverage external tools when their internal reasoning is insufficient. Therefore, uniformly penalizing tool invocation—whether through explicit tool-use costs or hard limits on the number of calls—may suppress useful exploration on genuinely challenging queries and undermine the very motivation of tool-augmented RL. The key question is thus not how to make models use fewer tools in general, but how to identify and penalize \emph{unnecessary} tool use: cases where the model could have solved the query without tools but nevertheless over-relies on them. R1-Searcher++ \cite{song2025r1searcherincentivizingdynamicknowledge} also recognizes this issue, but its penalty does not account for query difficulty. For instance, when only a small number of trajectories in a rollout group produce correct answers, imposing an overly strong tool-use penalty may hinder effective policy learning. In other words, a truly efficient reasoning model should learn not only how to use external tools, but also when to use them.

\begin{figure}
  \begin{center}
    \centerline{\includegraphics[width=\columnwidth]{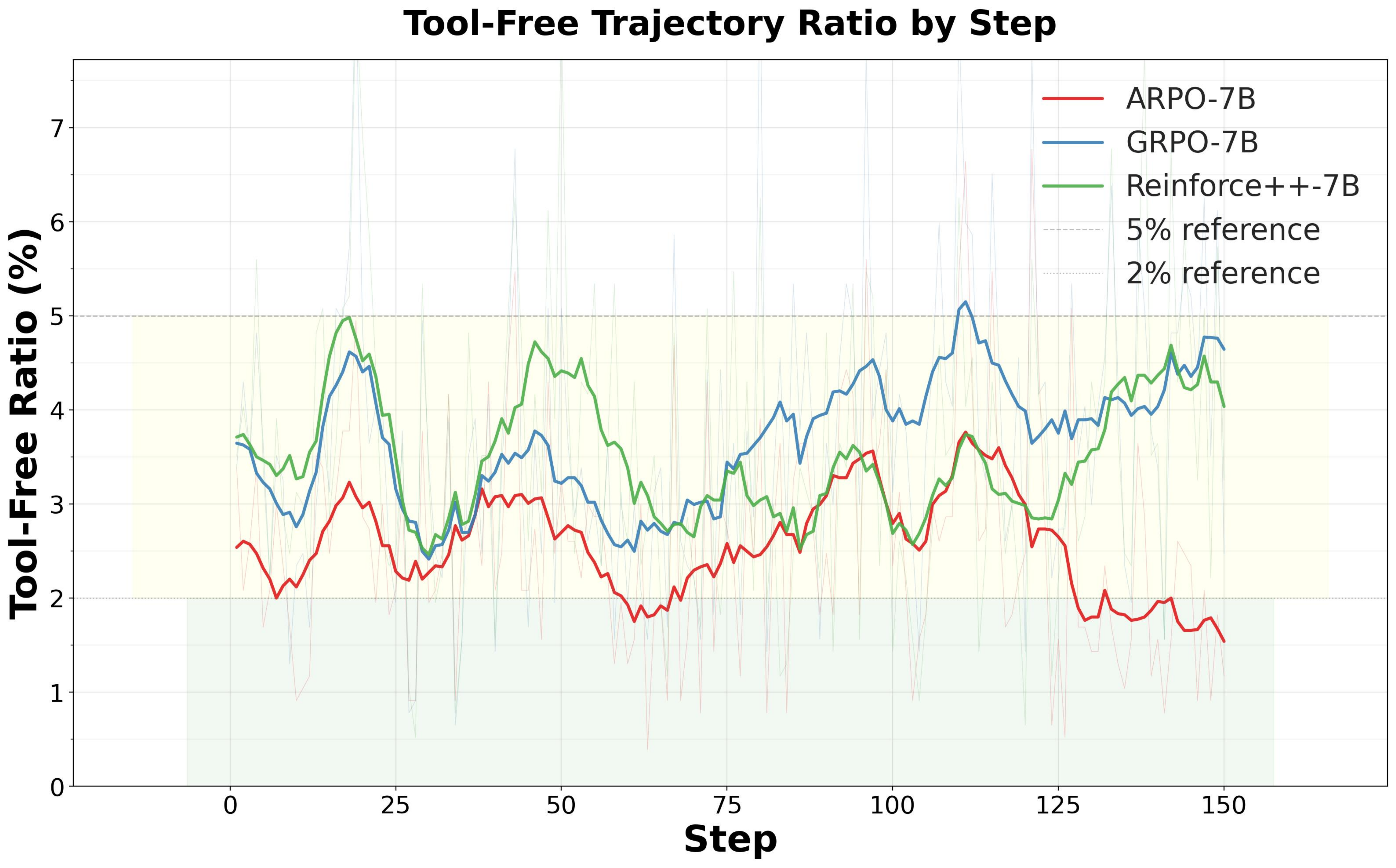}}
    \caption{
      Proportion of tool-free trajectories in rollouts across different algorithms.
    }
    \label{Fig2: rollout}
  \end{center}
  \vspace{-20pt}
\end{figure}

This distinction is difficult to capture in standard rollout procedures, where trajectories are usually sampled under tool-available settings and tool-free solutions may rarely appear. As illustrated in Figure~\ref{Fig2: rollout}, existing algorithms often produce very few trajectories that completely avoid tool invocation. Consequently, the training process lacks direct evidence about whether a query can be solved by the model's intrinsic reasoning ability alone. Without such evidence, it is difficult to distinguish necessary tool use from tool abuse.

To address this issue, we propose \textbf{E}fficient \textbf{A}gentic \textbf{P}olicy \textbf{O}ptimization (EAPO), a multi-tool agentic RL framework that optimizes efficiency without undermining tool-assisted reasoning. The core idea is to introduce \emph{tool-free rollouts}: for each query, EAPO ensures that the rollout set contains trajectories generated without external tools. These trajectories provide an explicit signal for whether the model can solve the query through internal reasoning alone. Based on this signal, EAPO applies difficulty-aware reward shaping that discourages tool use primarily when tool-free reasoning is sufficient, while preserving tool-assisted exploration for harder queries. In addition, EAPO incorporates confidence-aware token-level reweighting to improve policy learning by modulating gradients according to the model's generation confidence.

Empirically, EAPO improves both effectiveness and efficiency across nine reasoning benchmarks. Compared with GRPO, EAPO reduces tool usage by an average of 20\% on 7B-parameter models while improving task performance by approximately 10\%. These results suggest that agentic models can learn to use tools more selectively, avoiding unnecessary tool dependence without sacrificing the benefits of tool-augmented reasoning.

Our contributions are summarized as follows:
\begin{enumerate}
    \item We propose an efficiency-aware rollout mechanism that introduces tool-free trajectories, enabling the training process to distinguish necessary tool use from excessive tool dependence.
    \item We develop EAPO, a joint optimization framework that combines difficulty-aware reward shaping with confidence-aware token-level reweighting to improve both reasoning performance and tool-use efficiency.
    \item Extensive experiments on nine benchmarks across multiple model families (Qwen and Llama) and model scales (3B and 7B) demonstrate that EAPO consistently improves accuracy while substantially reducing unnecessary tool usage.
\end{enumerate}

%% file: Sections/2-preliminary.tex
\section{Preliminary}
\begin{figure*}[ht]
  \begin{center}
    \centerline{\includegraphics[width=\linewidth]{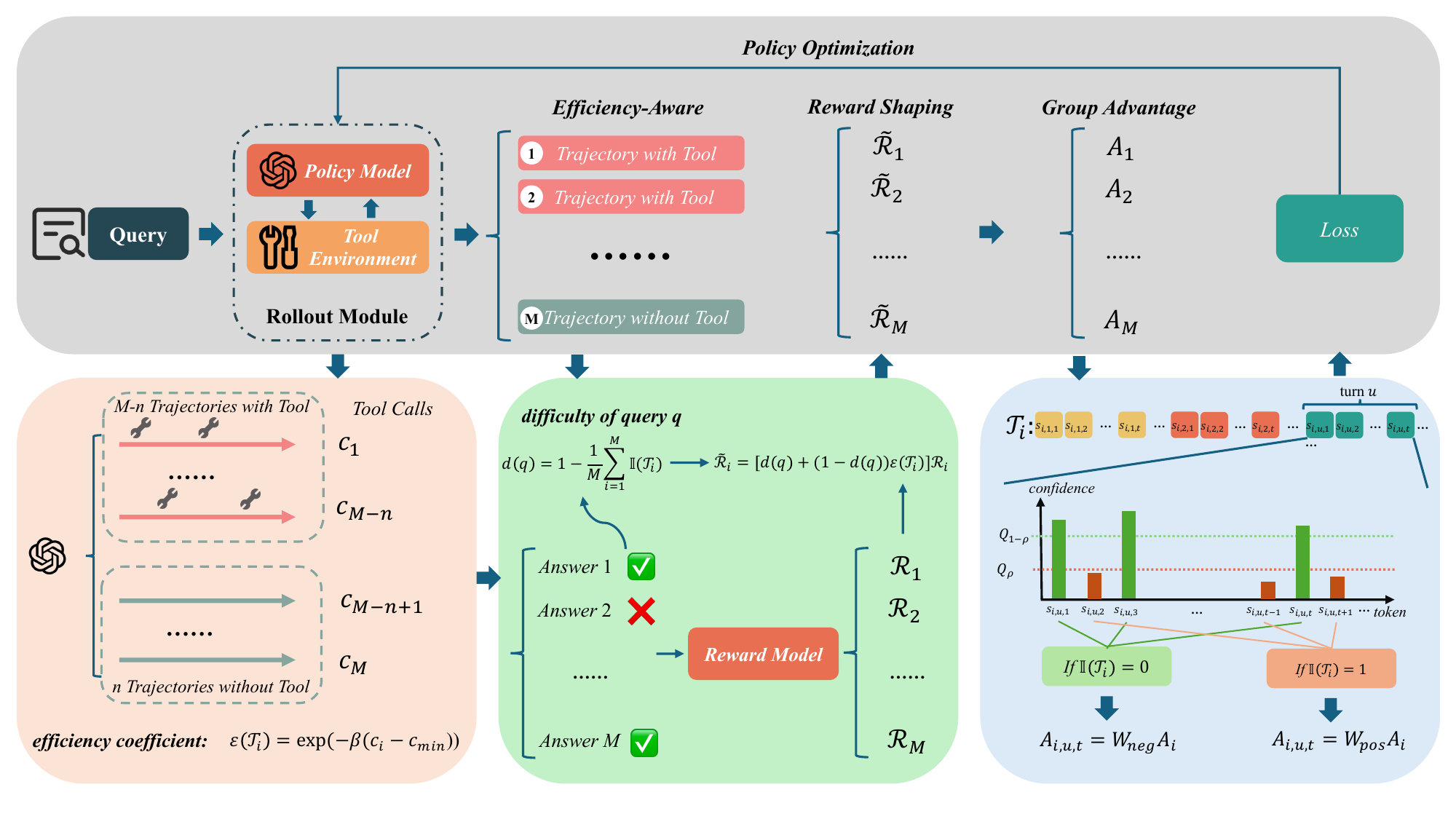}}
    \caption{
      The overview of EAPO algorithm.
    }
    \label{Fig3: EAPO}
  \end{center}
  \vspace{-20pt}
\end{figure*}
Before introducing EAPO, this section presents several concepts and formal problem definitions.

\subsection{Agentic Reinforcement Learning}
Following prior work \cite{dong2025arpo, dong2025aepo, xue2025simpletir}, we formulate the optimization objective of Agentic RL as:
\begin{equation}
\begin{aligned}
\max_{\pi_\theta} \;
\mathbb{E}_{x \sim \mathcal{D},\, y \sim \pi_\theta(\cdot \mid x; T)}
\Big[
r_\phi(x, y)
- \beta \, \mathcal{R}_{\mathrm{KL}}(\pi_\theta \mid x)
\Big],
\end{aligned}
\end{equation}
where \(T\) denotes the set of available tools, and \(\pi_\theta\) represents the policy LLM.
The reward function \(r_\phi(x, y)\) assigns a scalar feedback signal to each input–output pair \((x, y)\).
The input \(x\) is sampled from the dataset \(\mathcal{D}\), and \(y\) denotes the corresponding output sequence, which may include tool invocation results.
The regularization term \(\mathcal{R}_{\mathrm{KL}}(\pi_\theta \mid x)\) constrains the learned policy to remain close to a reference policy and is defined as:
\begin{equation}
\begin{aligned}
\mathcal{R}_{\mathrm{KL}}(\pi_\theta \mid x)
=
\mathbb{D}_{\mathrm{KL}}\!\left(
\pi_\theta(\cdot \mid x; T)
\,\big\|\,
\pi_{\mathrm{ref}}(\cdot \mid x; T)
\right),
\end{aligned}
\end{equation}
where \(\pi_{\mathrm{ref}}\) denotes the reference LLM.

\subsection{Token-level Confidence Estimation}
Consistent with prior work \cite{li2025knowexploredifficultyawarecertainty}, for the token $y_t$ generated at step $t$, we define the token-level confidence as the log-probability of the selected token under the policy distribution:
\begin{equation}
\label{eq:token_log_confidence}
C_t
\;=\;
\log p_t(y_t)
\;=\;
\log \pi_\theta\!\left(y_t \mid x, o_{<t}\right),
\end{equation}
where $p_t(\cdot)$ denotes the predictive distribution over the vocabulary at step $t$.
Since $0 < p_t(y_t) \leq 1$, the log-probability is upper-bounded by zero, and a larger value (closer to $0$) indicates higher model confidence in the generated token.

\subsection{Tool Abuse}
Building on existing work \cite{asai2023selfraglearningretrievegenerate, wang2025actingreasoningmoreteaching, song2025r1searcherincentivizingdynamicknowledge}, we define \textit{tool abuse} as follows: under a given task distribution and a tool set \(T\), if an agent policy \(\pi_\theta\) tends to generate output trajectories with redundant or excessive tool invocations, despite the existence of alternative solutions that achieve equivalent or near-equivalent performance (e.g., in terms of accuracy, reward, or utility), then the policy is said to exhibit tool abuse. For example, when the model is asked a trivially simple question such as "1+1=?", it should directly perform lightweight textual reasoning and provide an immediate response, rather than executing a series of Python programs or invoking calculator tools unnecessarily.

\subsection{Agentic Tool Design}
In this work, we primarily consider two types of tools:

(1) \textbf{Search tools}, which enable the model to retrieve external knowledge and mitigate hallucinations arising from insufficient or outdated internal knowledge; and

(2) \textbf{Python interpreters}, which allow the model to write and execute arbitrary Python code to assist in solving complex problems, such as those involving nontrivial numerical or symbolic computations.

%% file: Sections/3-method.tex
\section{Efficient Agentic Policy Optimization}

In this section, we introduce \textbf{E}fficient \textbf{A}gentic \textbf{P}olicy \textbf{O}ptimization (EAPO), an agentic reinforcement learning algorithm designed to mitigate agent tool abuse. As illustrated in Figure~\ref{Fig3: EAPO}, EAPO comprises three core components:
(1)~\textbf{Efficiency-Aware Rollout}, which enables the LLM to learn when tool usage is genuinely necessary by conducting dynamic rollouts and ensuring that, during advantage estimation, at least \(n\) trajectories do not invoke any tools;
(2)~\textbf{Difficulty-Aware Reward Shaping}, which accounts for the inherent difficulty of each query when computing rewards, thereby avoiding overly aggressive penalties on tool usage that could suppress exploration in challenging tasks;
and
(3)~\textbf{Confidence-Aware Advantage Reweighting}, which assigns different learning weights to tokens based on the model's confidence, reflecting their varying information density during loss computation. The full algorithm of EAPO is detailed
in Algorithm ~\ref{alg:eapo}.

\subsection{Efficiency-Aware Rollout}
Mitigating tool abuse requires models to learn \emph{when} tool use is necessary, rather than merely reducing tool calls. Prior work~\cite{wang2025actingreasoningmoreteaching, li2025torlscalingtoolintegratedrl} typically addresses this issue by imposing hard call limits or penalizing excessive tool usage. However, such uniform constraints do not distinguish queries solvable by intrinsic reasoning from those that truly require external tools. As a result, overly aggressive penalties may discourage tool use even on challenging queries, degrading performance.

To overcome this limitation, EAPO adopts a different strategy during training rollouts. Specifically, within each rollout consisting of \(M\) trajectories, we ensure that at least \(n\) trajectories do not invoke any tools. This is implemented by modifying the prompt to explicitly prohibit tool usage and by disabling tool execution in the rollout module when the model attempts to invoke tools. By explicitly contrasting tool-free and tool-augmented trajectories, the model can learn whether—and to what extent—tool usage yields genuine performance improvements for a given query. Let \(\mathcal{T}_i\) denote the \(i\)-th trajectory (\(i \in [1, M]\)). The efficiency coefficient \(\varepsilon(\mathcal{T}_i)\) for trajectory \(\mathcal{T}_i\) is defined as
\begin{equation}
    \varepsilon(\mathcal{T}_i) = \exp\!\bigl(-\beta (c_i - c_{\min})\bigr),
\end{equation}
where \(c_i\) denotes the number of tool calls in trajectory \(\mathcal{T}_i\), \(c_{\min}\) represents the minimum number of tool calls among the correct trajectories within the \(M\) sampled trajectories, and \(\beta\) is a hyperparameter controlling the strength of the tool-abuse penalty. If a trajectory does not involve any tool usage, we set \(c_i = c_{\min}\). In the absence of any correct trajectory, we set \(c_{\min} = 0\).

\subsection{Difficulty-Aware Reward Shaping}
The optimization objective of EAPO is to reduce unnecessary tool usage while ensuring that the model's performance does not degrade. Accordingly, during advantage estimation, it is crucial to account for whether the model has already mastered the given query. From the model's perspective, we define the difficulty of a query as follows:
\begin{equation}
d(q) \;=\;  1 - \frac{1}{M}\sum_{i=1}^{M}\mathbb{I}(\mathcal{T}_i),
\end{equation}

where $\mathbb{I}(\mathcal{T}_i)=1$ if trajectory $\mathcal{T}_i$ is correct, and $0$ otherwise. Taking the difficulty of the query \(q\) into account, we adjust the reward received by trajectory \(\mathcal{T}_i\) as follows:
\begin{equation}
\label{eq:reward_shaping}
\tilde{\mathcal{R}}_i
\;=\;
\Bigl[
d(q)
\;+\;
\bigl(1-d(q)\bigr)\,\varepsilon(\mathcal{T}_i)
\Bigr]\,
\mathcal{R}_i .
\end{equation}
Here, \(\mathcal{R}_i \) denotes the original reward of trajectory \(\mathcal{T}_i\). Following prior work \cite{dong2025arpo}, we adopt a hierarchical reward scheme that jointly considers answer correctness and the correctness of tool-usage formatting. The
overall reward R is formally defined as:
\begin{equation}
\mathcal{R}_i =
\begin{cases}
\mathrm{Acc.},
& \text{if Format is Good }, \\[6pt]
-1,
& \text{otherwise}.
\end{cases}
\end{equation}
Here, \(\mathrm{Acc.}\) denotes the model's answer accuracy: for mathematical tasks, it is a discrete value in \(\{0,1\}\), while for QA tasks it can be a continuous metric such as F1 or Recall.

\subsection{Confidence-Aware Advantage Reweighting}
\input{Tabs/main_math}
Inspired by prior work \cite{wang20258020rulehighentropyminority, tang2025rethinkingsamplepolarityreinforcement, li2025knowexploredifficultyawarecertainty}, tokens with different levels of confidence convey different amounts of information density. Accordingly, EAPO aims to emphasize less-confident tokens when the model produces correct answers, while down-weighting overconfident tokens when the model produces incorrect answers, thereby facilitating the learning of a more effective policy.

In general, in multi-turn tool usage, a turn refers to a single cycle in which the model performs reasoning, invokes a tool, observes the returned result, and then conducts further reasoning. That is, each turn typically contains one tool invocation. If a trajectory does not involve any tool usage, it is left unchanged and no additional processing is applied.

Given a trajectory $\mathcal{T}_i$ consisting of $U_i$ turns, we denote by
$\mathcal{S}_{i,u}=\{s_{i,u,1},\dots,s_{i,u,|\mathcal{S}_{i,u}|}\}$
the set of tokens in turn $u$ that participate in advantage computation,
after excluding tool-invocation segments (e.g., search query or code generation).
For each token $s_{i,u,t}\in\mathcal{S}_{i,u}$, we compute its confidence score
according to Equation~\ref{eq:token_log_confidence}.
Let $\{C_{i,u,t}\}_{t=1}^{|\mathcal{S}_{i,u}|}$ denote the resulting confidence values.

Let $\rho \in (0,1)$ be the reweighting ratio.
For each turn $u$, if the final answer of trajectory $\mathcal{T}_i$ is correct,
we assign an up-weighting factor $W_{\mathrm{pos}}$ to tokens whose confidence
values fall below the $\rho$-quantile of the empirical confidence distribution
within the same turn. Conversely, if the final answer is incorrect, we assign
a reweighting factor $W_{\mathrm{neg}}$ to tokens whose confidence values exceed
the $(1-\rho)$-quantile within that turn. Formally, the confidence-aware weight $\omega_{i,u,t}$ for token $s_{i,u,t}$ is defined as
\begin{equation}
\resizebox{\linewidth}{!}{$
\omega_{i,u,t} =
\begin{cases}
W_{\mathrm{pos}}, &
\text{if } \mathbb{I}(\mathcal{T}_i) = 1
\ \land\
C_{i,u,t} \le Q_{\rho}\!\Big(\{C_{i,u,k}\}_{k=1}^{|\mathcal{S}_{i,u}|}\Big), \\
W_{\mathrm{neg}}, &
\text{if } \mathbb{I}(\mathcal{T}_i) = 0
\ \land\
C_{i,u,t} \ge Q_{1-\rho}\!\Big(\{C_{i,u,k}\}_{k=1}^{|\mathcal{S}_{i,u}|}\Big), \\
1, & \text{otherwise},
\end{cases}
$}
\end{equation}
where $Q_{\rho}(\cdot)$ denotes the empirical $\rho$-quantile operator computed
over tokens within the same turn. Accordingly, EAPO redistributes the learning signal by assigning token-specific weights.
The reweighted advantage for token $s_{i,u,t}$ is defined as
\begin{equation}
A_{i,u,t} = \omega_{i,u,t}\, A_i,
\end{equation}
where $A_i$ is the trajectory-level advantage. The objective is to maximize the expected reweighted advantage over all tokens
participating in policy learning:
\begin{equation}
\begin{aligned}
\max_{\pi_\theta}\quad
& \mathbb{E}_{\mathcal{T}_i \sim \pi_\theta}
\Bigg[
\sum_{u=1}^{U_i}
\sum_{t \in \mathcal{S}_{i,u}}
A_{i,u,t} \\
& \qquad\qquad \cdot
\log \pi_\theta
\left(
s_{i,u,t}
\mid q, s_{i,u,<t}
\right)
\Bigg].
\end{aligned}
\end{equation}

%% file: Tabs/main_math.tex
\begin{table*}[htbp]
  \centering
  \scriptsize
  \setlength{\tabcolsep}{3pt}
  \captionsetup[table]{
    width=0.9\textwidth,
    justification=raggedright
  }
  \caption{Performance comparison of different methods on mathematical reasoning tasks. The best results are indicated in bold, and the second-best results are \underline{underlined}.}
  \label{tab:main_math}

  \resizebox{\textwidth}{!}{
  \begin{tabular}{@{}lcccccccccccc@{}}
    \toprule
    \multirow{2}{*}{Method} 
    & \multicolumn{2}{c}{\textbf{AIME24}} 
    & \multicolumn{2}{c}{\textbf{AIME25}} 
    & \multicolumn{2}{c}{\textbf{MATH500}} 
    & \multicolumn{2}{c}{\textbf{GSM8K}} 
    & \multicolumn{2}{c}{\textbf{MATH}}  
    & \multicolumn{2}{c}{\textbf{Avg.}} \\
    \cmidrule(lr){2-3} 
    \cmidrule(lr){4-5} 
    \cmidrule(lr){6-7} 
    \cmidrule(lr){8-9} 
    \cmidrule(lr){10-11} 
    \cmidrule(lr){12-13}
    & Pass@1 & TC & Pass@1 & TC & Pass@1 & TC & Pass@1 & TC & Pass@1 & TC & Pass@1 & TC \\
    \midrule

    \multicolumn{13}{c}{\textbf{\textit{Backbone Model: Qwen2.5-3B-Instruct}}} \\
    \midrule
    Base & 3.3 & - & 6.7 & - & 59.2 & - & 72.5 & - & 67.2 & - & 41.8 & -  \\
    TIR & 6.7 & 0.20 & 6.7 & 0.20 & 52.2 & 0.81 & 56.6 & 1.01 & 62.8 & 0.86 & 37.0 & 0.62 \\
    GRPO & \underline{13.3} & 1.10 & \underline{13.3} & 1.22 & \underline{70.0} & 1.14 & 81.6 & 1.13 & 80.5 & 1.07 & 51.7 & 1.13 \\
    Reinforce++ & \underline{13.3} & 1.21 & \textbf{16.7} & 1.30 & 69.4 & 1.09 & 84.8 & 1.16 & \textbf{81.8} & 1.12 & 53.2 & 1.18 \\
    ToolStar & \underline{13.3} & 1.03 & \underline{13.3} & 1.11 & 69.0 & 1.01 & 83.4 & 1.03 & 80.4 & 1.03 & 51.8 & 1.04 \\
    ARPO & \textbf{23.3} & 1.10 & \textbf{16.7} & 1.30 & 69.8 & 1.07 & \underline{85.0} & 1.04 & 80.8 & 1.05 & \textbf{55.1} & 1.12 \\
    \rowcolor{purple!10}
    EAPO(ours) & \textbf{23.3} & 0.90 & \underline{13.3} & 0.89 & \textbf{70.1} & 1.03 & \textbf{86.6} & 1.05 & \underline{81.6} & 1.01 & \underline{55.0} & 0.97 \\

    \midrule
    \multicolumn{13}{c}{\textbf{\textit{Backbone Model: Qwen2.5-7B-Instruct}}} \\
    \midrule
    Base & 10.0 & - & 10.0 & - & 70.6 & - & 90.2 & - & 82.0 & - & 52.6 & - \\
    TIR & 6.7 & 0.20 & 10.0 & 0.16 & 68.2 & 0.51 & 64.6 & 1.08 & 78.2 & 1.54 & 45.5 & 0.70 \\
    GRPO & 23.3 & 0.93 & 23.3 & 0.83 & 77.6 & 0.98 & 90.8 & 1.08 & 86.0 & 0.97 & 60.2 & 0.96 \\
    Reinforce++ & \underline{26.6} & 1.10 & 23.3 & 1.03 & 76.8 & 1.02 & 90.8 & 1.32 & 87.6 & 1.12 & 61.0 & 1.12 \\
    ToolStar & 23.3 & 1.86 & 20.0 & 1.40 & 76.8 & 1.38 & 90.8 & 1.08 & 86.4 & 1.28 & 59.4 & 1.32 \\
    ARPO & \underline{26.6} & 1.16 & \textbf{30.0} & 1.03 & \underline{78.0} & 1.02 & \underline{91.8} & 1.03 & 87.8 & 1.01 & \underline{62.8} & 1.05 \\
    AEPO & \textbf{30.0} & 1.20 & 23.3 & 1.21 & 77.6 & 1.10 & 91.4 & 1.01 & \underline{89.1} & 1.01 & 62.2 & 1.10 \\
    \rowcolor{purple!10}
    EAPO(ours) & \textbf{30.0} & 1.00 & \underline{26.6} & 0.93 & \textbf{78.8} & 0.98 & \textbf{92.0} & 1.01 & \textbf{90.0} & 1.03 & \textbf{63.5} & 0.99 \\

    \bottomrule
  \end{tabular}
  }
\end{table*}

%% file: Sections/4-experiments.tex
\section{Experiments}

\input{Tabs/main_qa}
\subsection{Datasets}
To comprehensively evaluate the effectiveness of the proposed EAPO, we conduct experiments on two major categories of reasoning tasks: (1) \textbf{Mathematical Reasoning:} This task covers AIME 2024 and AIME 2025, as well as widely used mathematical benchmarks including MATH500 \cite{lightman2023lets}, MATH \cite{hendrycksmath2021}, and GSM8K \cite{cobbe2021trainingverifierssolvemath}. (2) \textbf{Knowledge-Intensive Reasoning:} This task includes  HotpotQA \cite{yang2018hotpotqa}, 2WikiMultihopQA \cite{ho2020constructingmultihopqadataset}, Musique \cite{trivedi2021musique}, and Bamboogle \cite{press-etal-2023-measuring}.

\begin{figure*}[ht]
  \begin{center}
    \centerline{\includegraphics[width=\linewidth]{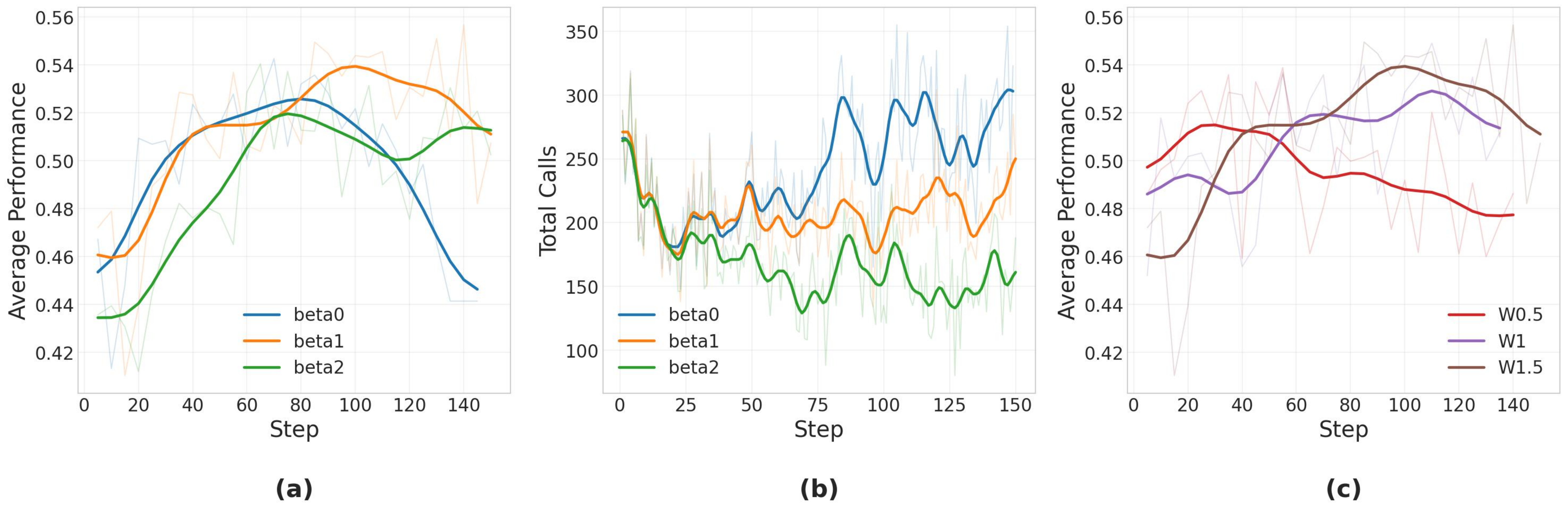}}
    \caption{
      Experimental results under different hyperparameter settings.
    }
    \label{Fig4: Hyperparameter}
  \end{center}
  \vspace{-20pt}
\end{figure*}

\subsection{Baseline}
To demonstrate the effectiveness of EAPO, we consider three categories of methods as baselines in our experiments: (1) \textbf{Training Free}: The model performs direct textual reasoning (Base), as well as tool-augmented reasoning guided by prompting (TIR);
(2) \textbf{Advanced RL Algorithms}:  GRPO \cite{shao2024deepseekmathpushinglimitsmathematical} and Reinforce++ \cite{hu2025reinforce++};(3) \textbf{Agentic RL Algorithms}: ToolStar \cite{dong2025tool}, ARPO \cite{dong2025arpo} and AEPO \cite{dong2025aepo}.

\subsection{Evaluation}
Following prior work \cite{dong2025aepo}, we adopt F1 and Exact Match (EM) as the evaluation metrics for knowledge-intensive reasoning tasks. For mathematical reasoning tasks, we employ the Qwen2.5-72B-Instruct model as an LLM-as-a-judge. During evaluation, the temperature is set to 0.6 and top-p to 0.95 by default, and performance is reported using the Pass@1 metric. Meanwhile, we use Tool Calls (TC) to denote the average number of tool invocations made by the model during inference.

\subsection{Implementation Details}
During EAPO training, we first follow prior work by constructing a 10K supervised fine-tuning (SFT) dataset for cold start, obtained by mixing AEPO \cite{dong2025aepo} search data with SimpleTIR \cite{xue2025simpletir} mathematical reasoning data. For the subsequent EAPO training stage, we use the same training data as AEPO . By default, the training batch size is set to 56, and the total number of rollouts is \(M = 16\), with at least \(n = 2\) trajectories guaranteed to involve no tool usage. In the main experiments, the hyperparameter \(\beta\) is fixed to 1, with \(W_{\text{pos}} = W_{\text{neg}} = 1.5\). The proportion of enhanced tokens is set to \(\rho = 0.2\).

To evaluate the effectiveness of EAPO, we adopt \textsc{Qwen2.5-3B-Instruct}, \textsc{Qwen2.5-7B-Instruct} and \textsc{Llama3.1-8B-Instruct} as backbone models, covering two representative model scales. For tool implementation, the search tool follows the Search-R1 setting \cite{jin2025searchr1}, with a locally deployed Wikipedia \cite{karpukhin-etal-2020-dense} retrieval service. We employ an E5-based retriever \cite{wang2024textembeddingsweaklysupervisedcontrastive} and, during both training and evaluation, retrieve the top-3 most relevant documents by default. The Python tool is implemented as a sandboxed execution environment that allows the model to safely execute code.

\subsection{Main Results}

As shown in Tables~\ref{tab:main_math} and~\ref{tab:main_qa} for Qwen models, and Tables~\ref{tab:main_math_llama} and~\ref{tab:main_qa_llama} for Llama models, EAPO consistently improves both effectiveness and efficiency across benchmarks and model scales.

\textbf{Effectiveness.} EAPO achieves the best or second-best performance on both knowledge-intensive and mathematical reasoning tasks, improving F1/EM and Pass@1 across datasets. These gains are stable across different model families and become more evident when scaling from 3B to 7B backbones, suggesting that EAPO benefits from increased model capacity. Compared with strong baselines such as GRPO, ARPO, and AEPO, EAPO attains higher overall accuracy without relying on excessive tool invocation, indicating that its improvements come from a better-calibrated agentic policy rather than brute-force tool use.

\textbf{Efficiency.} EAPO also maintains lower or comparable tool-call counts (TC) while achieving superior performance. In contrast, several baselines improve accuracy mainly by increasing tool usage, leading to a less favorable accuracy--efficiency trade-off. By encouraging tool-free trajectories and introducing difficulty-aware penalties, EAPO helps the model learn when tool invocation is truly necessary, thereby reducing redundant tool calls, especially on easier queries, without sacrificing accuracy.

Overall, these results demonstrate that EAPO achieves a stronger accuracy--efficiency trade-off than existing agentic RL methods, making it more suitable for practical deployment where tool use incurs non-trivial computational costs. Averaged across all nine benchmarks, using Pass@1 for mathematical reasoning and F1 for knowledge-intensive reasoning, EAPO improves performance over GRPO by 10.45\%, 7.27\%, and 9.69\% on Qwen2.5-3B, Qwen2.5-7B, and Llama3.1-8B, respectively, while reducing average tool calls by 18.33\%, 18.33\%, and 24.59\%.

\textbf{\textit{Finding 1.}} Simple prompt-based methods, such as few-shot prompting, are insufficient for effective multi-tool reasoning. Prompt-based TIR performs no better, and sometimes worse, than autonomous reasoning, consistent with prior findings~\cite{xue2025simpletir,dong2025aepo}.

\textbf{\textit{Finding 2.}} Efficiency gains are smaller on 3B models than on 7B models, as smaller models have less redundant tool use that can be reduced without hurting performance. This trend is expected to become stronger with larger models.
\begin{figure*}[ht]
  \begin{center}
    \centerline{\includegraphics[width=\linewidth]{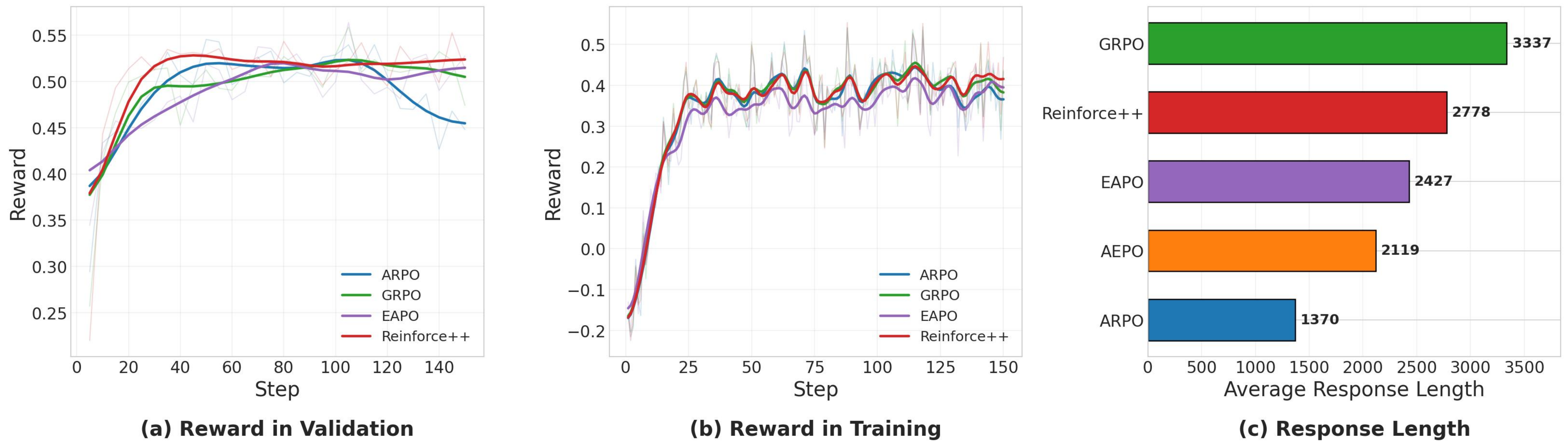}}
    \caption{
     Efficiency analysis of EAPO in training and inference.
    }
    \label{Fig5: Efficiency}
  \end{center}
  \vspace{-20pt}
\end{figure*}

\subsection{Hyperparameter Analysis}
We analyze the sensitivity of EAPO to key hyperparameters, with results shown in Figure~\ref{Fig4: Hyperparameter}.

(1) \textbf{Tool penalty parameter $\beta$}. As shown in Figure~\ref{Fig4: Hyperparameter}(a), a moderate tool-use penalty improves performance, whereas overly large penalties may over-constrain exploration and reduce accuracy. Figure~\ref{Fig4: Hyperparameter}(b) shows that different $\beta$ values yield similar performance during early training, roughly the first 25 steps. This is because the model is still relatively inaccurate, and the difficulty-aware coefficient in Equation~\ref{eq:reward_shaping} makes optimization less sensitive to tool penalties when accuracy is low. This confirms the role of difficulty-aware shaping in balancing effectiveness and efficiency.

Notably, the $\beta=0$ curve indicates that more tool calls do not necessarily improve performance. Tool use is beneficial only when it provides useful information for solving the current query; indiscriminate invocation may instead be harmful. Overall, EAPO is robust across a broad range of $\beta$, showing limited dependence on delicate tuning.

(2) \textbf{Confidence reweighting parameter $W$}. As shown in Figure~\ref{Fig4: Hyperparameter}(c), increasing the learning weights of informative tokens improves policy learning, consistent with prior findings~\cite{li2025knowexploredifficultyawarecertainty, wang20258020rulehighentropyminority}. In contrast, setting $W=0.5$ leads to clear degradation, suggesting that tokens with different information densities should not be treated uniformly. However, performance does not improve monotonically with larger $W$; effective reweighting should emphasize informative tokens without excessively distorting optimization.

\subsection{Detailed Efficiency Analysis}
In this section, we conduct a further analysis of efficiency. Specifically, we focus on the following two research questions:

(1)\textbf{ RQ1: Does the use of Efficiency-Aware Rollout in EAPO slow down model convergence?} As shown in Figure~\ref{Fig5: Efficiency} (a) and (b), under the same batch size and rollout number $M$, EAPO exhibits a convergence speed that is nearly identical to that of other baseline methods. The inclusion of a small number of tool-free trajectories in each rollout does not substantially hinder training efficiency or slow down convergence.

(2) \textbf{RQ2: Does using fewer tools in EAPO lead to increased token usage during inference, thereby reducing efficiency?} As illustrated in Figure~\ref{Fig5: Efficiency} (c), EAPO does not incur a significant increase in token usage compared to other approaches. On the contrary, EAPO consistently uses fewer tokens and fewer tool calls than methods such as GRPO and Reinforce++, resulting in more efficient and faster inference.

\input{Tabs/ablation}
\subsection{Ablation Study}
EAPO training includes three components: cold start, reward shaping with efficiency- and difficulty-aware coefficients, and advantage reweighting. We conduct ablations to assess each module, with results shown in Table~\ref{tab:ablation}.

The results show that removing cold start causes the largest performance drop, consistent across algorithms, indicating that supervised fine-tuning helps models acquire effective tool-use patterns. Without tool-abuse penalties, tool usage increases only mildly on math tasks but rises sharply on QA tasks, matching our main observations. Removing advantage reweighting makes the policy more conservative and less certain about when to use external tools.

%% file: Tabs/main_qa.tex
\begin{table*}[htbp]
  \centering
  \scriptsize
  \setlength{\tabcolsep}{2.2pt}
  \captionsetup[table]{
    width=0.9\textwidth,
    justification=raggedright
  }
  \caption{Performance comparison of different methods on knowledge-intensive reasoning tasks. The best results are indicated in bold, and the second-best results are \underline{underlined}.}
  \label{tab:main_qa}

  \resizebox{\textwidth}{!}{
  \begin{tabular}{@{}lccccccccccccccc@{}}
    \toprule
    \multirow{2}{*}{Method} 
    & \multicolumn{3}{c}{\textbf{HotpotQA}} 
    & \multicolumn{3}{c}{\textbf{2Wiki}} 
    & \multicolumn{3}{c}{\textbf{MuSiQue}} 
    & \multicolumn{3}{c}{\textbf{Bamboogle}}  
    & \multicolumn{3}{c}{\textbf{Avg.}} \\
    \cmidrule(lr){2-4} 
    \cmidrule(lr){5-7} 
    \cmidrule(lr){8-10} 
    \cmidrule(lr){11-13} 
    \cmidrule(lr){14-16} 
    & F1 & EM & TC 
    & F1 & EM & TC 
    & F1 & EM & TC 
    & F1 & EM & TC 
    & F1 & EM & TC \\
    \midrule

    \multicolumn{16}{c}{\textbf{\textit{Backbone Model: Qwen2.5-3B-Instruct}}} \\
    \midrule
    Base & 9.7 & 7.2 & - & 9.4 & 6.5 & - & 3.6 & 0.5 & - & 11.7 & 7.4 & - & 8.6 & 5.4 & - \\
    TIR & 15.4 & 10.4 & 1.44 & 14.1 & 9.4 & 1.60 & 6.1 & 3.2 & 1.59 & 16.4 & 10.0 & 1.32 & 13.0 & 8.3 & 1.49 \\
    GRPO & 40.7 & 31.8 & 2.70 & 42.1 & 38.9 & 2.41 & 17.7 & 9.8 & 2.48 & 44.8 & 34.5 & 2.57 & 36.3 & 28.8 & 2.54 \\
    Reinforce++ & \underline{42.3} & \underline{37.5} & 2.99 & 39.7 & 37.4 & 2.62 & 16.0 & 7.8 & 2.51 & 39.4 & 30.6 & 2.27 & 34.4 & 28.3 & 2.60 \\
    ToolStar & 38.5 & 30.0 & 2.29 & 31.5 & 27.5 & 2.72 & 15.7 & 8.5 & 2.52 & 40.7 & 30.4 & 2.31 & 31.6 & 24.1 & 2.46 \\
    ARPO & 41.2 & 32.5 & 2.28 & \underline{45.4} & \underline{40.5} & 2.37 & \underline{19.4} & \underline{11.5} & 2.62 & \textbf{48.1} & \underline{38.4} & 2.13 & \underline{38.5} & \underline{30.7} & 2.35 \\
    \rowcolor{purple!10}
    EAPO(ours) & \textbf{46.5} & \textbf{38.5} & 1.84 & \textbf{53.4} & \textbf{45.5} & 2.04 & \textbf{23.4} & \textbf{17.5} & 2.26 & \underline{48.0} & \textbf{40.0} & 1.90 & \textbf{42.9} & \textbf{35.4} & 2.01 \\

    \midrule
    \multicolumn{16}{c}{\textbf{\textit{Backbone Model: Qwen2.5-7B-Instruct}}} \\
    \midrule
    Base & 10.6 & 5.6 & - & 11.6 & 5.0 & - & 6.5 & 2.5 & - & 20.1 & 12.0 & - & 12.2 & 6.3 & - \\
    TIR & 13.8 & 6.7 & 2.37 & 12.1 & 5.1 & 2.91 & 7.7 & 2.5 & 2.56 & 11.2 & 7.3 & 2.30 & 11.2 & 5.4 & 2.54 \\
    GRPO & \textbf{54.4} & \textbf{42.5} & 2.62 & 45.0 & 37.1 & 2.95 & \underline{32.1} & \underline{17.5} & 3.10 & \underline{55.6} & 44.8 & 2.80 & \underline{46.8} & \underline{35.5} & 2.87 \\
    Reinforce++ & 50.6 & 40.0 & 1.81 & 44.2 & 36.5 & 2.09 & 30.7 & 15.4 & 2.26 & 54.2 & 42.4 & 1.94 & 44.9 & 33.6 & 2.03 \\
    ToolStar & 42.4 & 32.5 & 3.82 & 42.1 & 30.9 & 4.89 & 16.8 & 11.0 & 4.07 & 48.0 & 36.0 & 4.28 & 37.3 & 27.6 & 4.27 \\
    ARPO & 45.7 & 34.5 & 1.93 & 44.9 & 40.0 & 2.02 & 22.5 & 12.0 & 2.33 & 51.8 & 40.0 & 1.93 & 41.2 & 31.6 & 2.05 \\
    AEPO & 47.7 & 36.0 & 4.26 & \underline{48.8} & \underline{42.5} & 4.95 & 25.1 & 13.5 & 4.71 & \underline{55.6} & \underline{46.4} & 4.16 & 44.3 & 34.6 & 4.52 \\
    \rowcolor{purple!10}
    EAPO(ours) & \underline{54.1} & \textbf{42.5} & 1.99 & \textbf{58.6} & \textbf{51.5} & 1.71 & \textbf{33.1} & \textbf{22.0} & 2.46 & \textbf{60.4} & \textbf{48.8} & 2.17 & \textbf{51.6} & \textbf{41.2} & 2.08 \\

    \bottomrule
  \end{tabular}
  }
\end{table*}

%% file: Tabs/ablation.tex
\begin{table}[htbp]
  \centering
  \setlength{\tabcolsep}{4pt}
  \caption{Ablation Study Results.}
  \label{tab:ablation}
\resizebox*{0.95 \linewidth}{!}{
  \begin{tabular}{@{}c ccc cc@{}}
    \toprule
    & \multicolumn{3}{c}{\textbf{HotpotQA}} & \multicolumn{2}{c}{\textbf{MATH}} \\
    \cmidrule(lr){2-4} \cmidrule(lr){5-6}
    Method & F1 & EM & TC & Pass@1 & TC \\
    \midrule
    EAPO & 54.1 & 42.5 & 1.99 & 90.0 & 1.03 \\
    \midrule
    w/o SFT & 32.7 & 28.6 & 2.03 & 75.4& 1.12 \\
    w/o Reward Shaping & 51.2 & 39.0 & 3.49 & 87.6 & 1.06 \\
    w/o Advantage Reweight & 51.4 & 41.5 & 1.48 & 86.6 & 0.99 \\
    \bottomrule
  \end{tabular}
}
\end{table}

%% file: Sections/5-appendix.tex
\appendix

\section{Related Works}
\paragraph{Agentic Reasoning and Tool-Integrated LLMs.}
Early work such as \textbf{ReAct}~\cite{yao2022react} establishes a foundational paradigm for agentic reasoning by tightly interleaving natural language reasoning with external actions. By enabling models to ``think while acting,'' ReAct demonstrates the feasibility and effectiveness of integrating tool interactions into the reasoning process, laying the conceptual groundwork for subsequent agentic reinforcement learning approaches. However, ReAct primarily focuses on expressivity and task solvability, without explicitly addressing efficiency or optimal tool usage.

\paragraph{Agentic Reinforcement Learning with Tool Execution.}
Recent agentic RL methods, including \textbf{ToRL}~\cite{li2025torlscalingtoolintegratedrl}, \textbf{SimpleTIR}~\cite{xue2025simpletir}, and \textbf{Search-R1}~\cite{jin2025searchr1}, leverage reinforcement learning to train LLMs to invoke tools (e.g., Python execution or retrieval) for mathematical and reasoning tasks. These approaches typically adopt outcome-based reward designs centered on final answer correctness, and control tool abuse by imposing a hard upper bound on the number of tool calls per rollout. While effective in restricting excessive tool usage, such rigid constraints introduce two key limitations: (1) for complex problems that inherently require multi-step tool-based reasoning, a small fixed tool-call budget can prematurely truncate valid trajectories; and (2) forced termination upon reaching the tool limit results in truncated rollouts and discontinuous reward signals, complicating long-horizon credit assignment and potentially destabilizing training.

\paragraph{Explicit Optimization of Tool-Use Efficiency.}
To directly address tool overuse, \textbf{OTC}~\cite{wang2025actingreasoningmoreteaching} proposes a tool-integrated reward that jointly accounts for solution correctness and tool-call efficiency. By explicitly penalizing excessive tool usage, OTC encourages models to reduce unnecessary external interactions while maintaining competitive performance, and demonstrates its effectiveness under standard RL algorithms such as PPO and GRPO. However, in tool-integrated reinforcement learning, introducing strong tool penalties early in training may discourage exploration of tool usage, causing policies to converge toward tool-avoidant behaviors rather than learning when tool invocation is genuinely beneficial. Moreover, aggressive efficiency penalties can introduce additional instability into the optimization process and potentially degrade final performance.

\paragraph{Retrieval-Centric and Knowledge-Internalization Approaches.} 
\textbf{R1-Searcher++} adopts a two-stage framework that combines supervised fine-tuning and reinforcement learning to train LLMs that dynamically integrate internal reasoning with external search ~\cite{song2025r1searcherincentivizingdynamicknowledge}. By encouraging the internalization of retrieved knowledge, the method reduces redundant retrieval over time. Nevertheless, it does not explicitly regulate tool-call efficiency at the policy level and assumes that external information can be effectively absorbed into model parameters, limiting its applicability to procedural or computation-oriented tools and offering limited control over tool usage behavior during rollout.

\paragraph{Efficiency-Aware Reinforcement Learning via Length Regularization.}
Several works explore efficiency from the perspective of generation length. \textbf{GRPO-LEAD}~\cite{zhang2025grpolead} and \textbf{Stable Reinforcement Learning for Efficient Reasoning}~\cite{dai2025stablerl} introduce length-aware reward designs that favor concise correct outputs, and further analyze the instability induced by naive length penalties. While effective in reducing verbosity, these methods reveal a key limitation: uniform length penalties often harm performance on harder problems that inherently require longer reasoning chains. This observation suggests that efficiency regularization should be adaptive and difficulty-aware, rather than uniformly applied across all queries.

\section{Implementation Details}
\subsection{Datasets}
\subsubsection{Mathematical Reasoning}
\begin{itemize}
    \item \textbf{AIME2024} is a dataset that contains 30 math problems from the 2024 American Invitational Mathematics Examination (AIME I and II) along with their solutions, answers, source URLs, and year labels. It’s provided as a single training split for evaluating mathematical reasoning and problem-solving in models using real competition questions.
    \item \textbf{AIME2025} is a dataset of 30 problems and answers from the 2025 American Invitational Mathematics Examination (AIME 2025) formatted with fields for the problem text, the correct integer answer, and an identifier. It is provided as a single split (test) with real competition questions designed to benchmark mathematical reasoning and problem-solving in models.
    \item \textbf{MATH} \cite{hendrycksmath2021} is a benchmark of challenging high-school competition math problems (with step-by-step solutions) designed to assess and improve mathematical problem-solving in machine learning models. It spans various subjects and difficulty levels, enabling fine-grained evaluation of reasoning, and highlights the gap between current models and human performance.
    \item \textbf{GSM8K} is a benchmark of human-written grade-school math word problems designed to evaluate multi-step numerical reasoning in language models. Problems require a small number of basic arithmetic operations and emphasize reasoning and language understanding over pattern memorization.

    \item \textbf{MATH500} is a benchmark for advanced mathematical reasoning, consisting of challenging problems with step-by-step solutions and exact final answers. It covers multiple mathematical domains and difficulty levels, focusing on multi-step symbolic reasoning and precise evaluation.
\end{itemize}

\subsubsection{Knowledge-Intensive Reasoning}
\begin{itemize}
\item \textbf{HotpotQA} \cite{yang2018hotpotqa} is a multi-hop question answering dataset based on Wikipedia that targets complex reasoning across multiple documents. Questions are diverse and not restricted to fixed templates. The dataset provides sentence-level supporting facts, enabling evaluation of both answer accuracy and explainable reasoning, and includes comparison questions that further increase reasoning difficulty.

\item \textbf{2WikiMultiHopQA} \cite{ho2020constructingmultihopqadataset} focuses on multi-hop question answering that requires integrating evidence from multiple Wikipedia passages. Each example includes a question and annotated supporting evidence, allowing evaluation of answer correctness as well as evidence identification, with an emphasis on cross-document reasoning rather than single-paragraph retrieval.

\item \textbf{MuSiQue} \cite{trivedi2021musique} is a multi-hop QA dataset constructed by composing related single-hop questions into connected reasoning chains. Answering later steps depends on earlier ones, encouraging genuine multi-step inference and reducing shortcut solutions. The dataset also includes contrast unanswerable questions to further challenge models’ reasoning and evidence integration abilities.

\item  \textbf{Bamboogle} \cite{press-etal-2023-measuring} Bamboogle is a small, manually constructed multi-hop QA dataset with 2-hop questions written by authors such that answers cannot be reliably found using a popular search engine, but evidence for both reasoning steps exists in Wikipedia. It is designed to evaluate a model’s ability to perform varied compositional reasoning across independent questions beyond templated examples.
\end{itemize}
\subsection{Baseline}
\subsubsection{Base Model}
\begin{itemize}
    \item \textbf{Qwen2.5} \cite{qwen2024qwen25} is a new generation of large language models that significantly improves general language understanding, reasoning, coding, and long-context capabilities. It is trained on a much larger and higher-quality corpus and further enhanced through large-scale supervised fine-tuning and reinforcement learning. The series includes models of various sizes and demonstrates strong performance across a wide range of benchmarks, serving as a foundation for multiple specialized variants.
    \item \textbf{Llama3.1} \cite{grattafiori2024llama3herdmodels} is an advanced open-weight large language model series developed by Meta, featuring improved multilingual understanding, reasoning, tool-use capability, and long-context modeling. It is trained on large-scale high-quality data and further aligned through supervised fine-tuning and preference optimization. The series includes models at multiple scales and exhibits strong performance across diverse downstream tasks, making it a widely adopted foundation for both general-purpose and agentic reasoning research.
\end{itemize}
\input{Tabs/res_llama_math}
\input{Tabs/res_llama_qa}
\subsubsection{Advanced RL Algorithms}
\begin{itemize}
    \item \textbf{GRPO} \cite{shao2024deepseekmathpushinglimitsmathematical} is a reinforcement learning algorithm based on policy optimization that aims to balance training stability, sample efficiency, and theoretical policy improvement guarantees, which are often difficult to achieve simultaneously in traditional policy optimization methods. GRPO introduces the concept of relative advantage, enabling a simplified and more stable advantage estimation while preserving the theoretical guarantee of monotonic policy improvement. This design reduces variance and computational overhead compared to conventional advantage estimators. As a result, GRPO is applicable to reinforcement learning tasks with both continuous and discrete action spaces, and is particularly suitable for large-scale policy learning scenarios.
    \item \textbf{Reinforce++} \cite{hu2025reinforce++} is an enhanced variant of the classic REINFORCE algorithm, designed to overcome its inherent limitations and improve training efficiency and stability. It integrates multiple optimization strategies, including the use of a baseline function to reduce gradient variance through baseline subtraction. By combining baseline estimation with temporal-difference (TD) learning, REINFORCE++ enables more stable and low-variance gradient updates and supports incremental updates without waiting for complete trajectories. Furthermore, the algorithm employs entropy regularization to encourage exploration and prevent premature policy convergence, thereby improving robustness and overall learning performance.
    \item \textbf{ToolStar} \cite{dong2025tool} is a reinforcement learning (RL) framework that trains large language models (LLMs) to autonomously use and coordinate multiple external tools during step-by-step reasoning. It includes a scalable pipeline for synthesizing tool-use data and a two-stage training approach (cold-start supervised fine-tuning followed by multi-tool self-critic RL with hierarchical rewards) to improve collaborative tool invocation. Experimental results show that Tool-Star enhances both reasoning accuracy and effective tool usage across diverse reasoning benchmarks
    \item \textbf{ARPO} \cite{dong2025arpo} is a reinforcement learning algorithm designed to train multi-turn language model agents that interact with external tools more effectively. It introduces an entropy-based adaptive rollout mechanism that increases exploration when the model is uncertain after making tool calls, and an advantage attribution estimation to properly credit shared and branching reasoning paths. This approach improves multi-step tool use and reasoning performance on challenging benchmarks while reducing training complexity and tool-call requirements compared to traditional RL methods.
    \item \textbf{AEPO} \cite{dong2025aepo} is a reinforcement learning algorithm that balances entropy during both rollout sampling and policy updates to improve stability and performance when training multi-turn web-capable agents. It introduces a dynamic entropy-balanced rollout mechanism to manage branching exploration and an entropy-aware policy update to preserve useful gradients on uncertain decisions. AEPO outperforms several existing RL baselines on diverse reasoning and tool-use benchmarks, enabling more efficient and robust agentic behavior.

\end{itemize}

\section{AI Assistant Use}
We used ChatGPT to polish the language of the paper.

\begin{algorithm}[t]
\caption{Efficient Agentic Policy Optimization}
\label{alg:eapo}
\begin{algorithmic}[1]
\Require Policy $\pi_\theta$, reference policy $\pi_{\mathrm{ref}}$, dataset $\mathcal{D}$, tools $\mathcal{T}$, rollout size $M$, tool-free size $n$, penalty $\beta$, ratio $\rho$, weights $W_{\mathrm{pos}}, W_{\mathrm{neg}}$
\Ensure Optimized policy $\pi_\theta$

\For{each training iteration}
    \State Sample mini-batch $\mathcal{D}_b \subset \mathcal{D}$
    \For{each query $q \in \mathcal{D}_b$}
        \Statex \textcolor{blue}{// Efficiency-Aware Rollout}
        \State Generate $n$ tool-free trajectories by disabling tools
        \State Generate $M-n$ tool-available trajectories with $\mathcal{T}$
        \State Let $\mathcal{P}=\{\mathcal{T}_i\}_{i=1}^{M}$

        \Statex \textcolor{blue}{// Difficulty-Aware Reward Shaping}
        \For{each $\mathcal{T}_i \in \mathcal{P}$}
            \State Compute correctness $\mathbb{I}(\mathcal{T}_i)$, tool calls $c_i$, and base reward $R_i$
        \EndFor
        \State $d(q) \gets 1-\frac{1}{M}\sum_{i=1}^{M}\mathbb{I}(\mathcal{T}_i)$
        \State $c_{\min} \gets \min_{\mathbb{I}(\mathcal{T}_j)=1} c_j$, or $0$ if no correct trajectory
        \For{each $\mathcal{T}_i \in \mathcal{P}$}
            \State $\hat{c}_i \gets c_{\min}$ if $\mathcal{T}_i$ is tool-free, otherwise $c_i$
            \State $\varepsilon_i \gets \exp[-\beta(\hat{c}_i-c_{\min})]$
            \State $\widetilde{R}_i \gets [d(q)+(1-d(q))\varepsilon_i]R_i$
        \EndFor
        \State Compute trajectory advantage $A_i$ from $\{\widetilde{R}_i\}_{i=1}^{M}$

        \Statex \textcolor{blue}{// Confidence-Aware Advantage Reweighting}
        \For{each token $s_{i,u,t}$ used for policy learning}
            \State $C_{i,u,t}\gets \log\pi_\theta(s_{i,u,t}\mid q,s_{i,u,<t})$
            \State Set $\omega_{i,u,t}=W_{\mathrm{pos}}$ for correct low-confidence tokens
            \State Set $\omega_{i,u,t}=W_{\mathrm{neg}}$ for incorrect high-confidence tokens
            \State Otherwise set $\omega_{i,u,t}=1$
            \State $A_{i,u,t}\gets \omega_{i,u,t}A_i$
        \EndFor

        \Statex \textcolor{blue}{// Policy Optimization}
        \State Update $\pi_\theta$ with reweighted policy-gradient objective and KL regularization to $\pi_{\mathrm{ref}}$
    \EndFor
\EndFor
\State \Return $\pi_\theta$
\end{algorithmic}
\end{algorithm}

%% file: Tabs/res_llama_math.tex
\begin{table*}[htbp]
  \centering
  \scriptsize
  \setlength{\tabcolsep}{3pt}
  \captionsetup[table]{
    width=0.9\textwidth,
    justification=raggedright
  }
  \caption{Performance comparison of different methods on mathematical reasoning tasks using Llama3.1-8B-Instruct as the backbone. The best results are indicated in bold, and the second-best results are \underline{underlined}.}
  \label{tab:main_math_llama}

  \resizebox{\textwidth}{!}{
  \begin{tabular}{@{}lcccccccccccc@{}}
    \toprule
    \multirow{2}{*}{Method} 
    & \multicolumn{2}{c}{\textbf{AIME24}} 
    & \multicolumn{2}{c}{\textbf{AIME25}} 
    & \multicolumn{2}{c}{\textbf{MATH500}} 
    & \multicolumn{2}{c}{\textbf{GSM8K}} 
    & \multicolumn{2}{c}{\textbf{MATH}}  
    & \multicolumn{2}{c}{\textbf{Avg.}} \\
    \cmidrule(lr){2-3} 
    \cmidrule(lr){4-5} 
    \cmidrule(lr){6-7} 
    \cmidrule(lr){8-9} 
    \cmidrule(lr){10-11} 
    \cmidrule(lr){12-13}
    & Pass@1 & TC 
    & Pass@1 & TC 
    & Pass@1 & TC  
    & Pass@1 & TC 
    & Pass@1 & TC 
    & Pass@1 & TC \\
    \midrule

    Base 
    & 0.0 & - 
    & 3.3 & - 
    & 47.2 & - 
    & 79.6 & - 
    & 62.4 & - 
    & 38.50 & - \\

    TIR 
    & 3.3 & 0.52 
    & 3.3 & 0.73 
    & 45.3 & 0.79 
    & 70.79 & 1.22 
    & 58.0 & 0.97 
    & 36.14 & 0.85 \\

    GRPO 
    & 10.0 & 1.02 
    & 13.3 & 1.27 
    & 58.8 & 1.08 
    & \underline{86.2} & 1.11 
    & \underline{78.4} & 1.08 
    & 49.34 & 1.11 \\

    Reinforce++ 
    & \underline{13.3} & 1.08 
    & \underline{16.7} & 1.10 
    & \underline{59.7} & 1.06 
    & 80.6 & 1.48 
    & 76.8 & 1.20 
    & \underline{49.42} & 1.18 \\

    \rowcolor{purple!10}
    EAPO(ours) 
    & \textbf{26.7} & 1.00 
    & \textbf{20.0} & 0.96 
    & \textbf{62.7} & 0.98 
    & \textbf{86.8} & 0.96 
    & \textbf{79.2} & 1.03 
    & \textbf{55.08} & 0.99 \\

    \bottomrule
  \end{tabular}
  }
\end{table*}

%% file: Tabs/res_llama_qa.tex
\begin{table*}[htbp]
  \centering
  \scriptsize
  \setlength{\tabcolsep}{2.2pt}
  \captionsetup[table]{
    width=0.9\textwidth,
    justification=raggedright
  }
  \caption{Performance comparison of different methods on knowledge-intensive reasoning tasks using Llama3.1-8B-Instruct as the backbone. The best results are indicated in bold, and the second-best results are \underline{underlined}.}
  \label{tab:main_qa_llama}

  \resizebox{\textwidth}{!}{
  \begin{tabular}{@{}lccccccccccccccc@{}}
    \toprule
    \multirow{2}{*}{Method} 
    & \multicolumn{3}{c}{\textbf{HotpotQA}} 
    & \multicolumn{3}{c}{\textbf{2Wiki}} 
    & \multicolumn{3}{c}{\textbf{MuSiQue}} 
    & \multicolumn{3}{c}{\textbf{Bamboogle}}  
    & \multicolumn{3}{c}{\textbf{Avg.}} \\
    \cmidrule(lr){2-4} 
    \cmidrule(lr){5-7} 
    \cmidrule(lr){8-10} 
    \cmidrule(lr){11-13} 
    \cmidrule(lr){14-16}
    & F1 & EM & TC 
    & F1 & EM & TC 
    & F1 & EM & TC 
    & F1 & EM & TC 
    & F1 & EM & TC \\
    \midrule

    Base 
    & 22.3 & 7.2 & - 
    & 23.7 & 10.2 & - 
    & 9.8 & 7.4 & - 
    & 26.7 & 12.5 & - 
    & 20.6 & 9.3 & - \\

    TIR 
    & 30.7 & 29.8 & 3.12 
    & 33.7 & 28.6 & 2.98 
    & 17.8 & 9.7 & 2.67 
    & 30.6 & 15.8 & 2.42 
    & 28.2 & 20.9 & 2.80 \\

    GRPO 
    & 58.8 & \underline{44.3} & 2.89 
    & \underline{48.0} & \underline{40.7} & 3.27 
    & \underline{30.6} & \textbf{16.7} & 3.07 
    & 57.4 & \underline{42.8} & 2.94 
    & 48.7 & \underline{36.1} & 3.04 \\

    Reinforce++ 
    & \underline{59.2} & 43.9 & 1.97 
    & 47.1 & 38.4 & 2.13 
    & 30.2 & 15.9 & 2.37 
    & \underline{58.4} & 23.7 & 2.20 
    & \underline{48.7} & 30.4 & 2.17 \\

    \rowcolor{purple!10}
    EAPO(ours) 
    & \textbf{62.4} & \textbf{46.7} & 1.98 
    & \textbf{52.0} & \textbf{44.8} & 1.69 
    & \textbf{32.8} & \underline{16.0} & 2.58 
    & \textbf{61.7} & \textbf{47.9} & 2.19 
    & \textbf{52.2} & \textbf{38.8} & 2.11 \\

    \bottomrule
  \end{tabular}
  }
\end{table*}